\documentclass[runningheads]{llncs}
\usepackage[T1]{fontenc}
\usepackage{graphicx,verbatim}
\usepackage{tabularx}
\usepackage{mathtools}
\usepackage[table,xcdraw]{xcolor}
\usepackage[table]{xcolor}
\usepackage{multirow}
\usepackage{hyperref}
\usepackage{colortbl}
\usepackage{xcolor}
\usepackage{amssymb}
\definecolor{nodulecolor}{RGB}{255,77,0}
\usepackage[utf8]{inputenc}

\begin{document}
\title{Predicting Radiologist Expertise from 3D Gaze Patterns During CT Interpretation}
\titlerunning{Predicting Radiologist Expertise from 3D Gaze Patterns}

\author{Leila Khaertdinova\inst{1,*}
\and
Anna Anikina\inst{1,*}
\and
Claudia Mello-Thoms\inst{2}
\and
Bulat Ibragimov\inst{1}
}
%
\authorrunning{L. Khaertdinova, A. Anikina et al.}

\institute{
  Department of Computer Science, University of Copenhagen, Copenhagen,~Denmark \\
  \email{\{leila.khaertdinova, anan, bulat\}@di.ku.dk} \and
  Department of Radiology, University of Iowa, Iowa, United States \\
  \email{claudia-mello-thoms@uiowa.edu}
}
%
\def\thefootnote{*}
\footnotetext{These authors contributed equally to this work.}

\maketitle              
\begin{abstract}

Accurate interpretation of volumetric CT requires efficient navigation of 3D image volumes and attention to diagnostically relevant regions. While eye-tracking has been widely studied in 2D medical imaging, its use for expertise assessment in CT settings remains limited. We propose a gaze-informed transformer framework for expertise classification in thoracic CT. Using a DINOv2 backbone, radiologist fixation patterns are integrated into volumetric feature learning through (1) a learnable log-space bias in self-attention and (2) gaze-weighted pooling of patch embeddings. We trained and evaluated our approach on 182 CT reading sessions from five radiologists with varying levels of experience. On a held-out test set, the model achieves an ROC-AUC of 0.91 and F1 score of 0.86, outperforming adapted methods. These findings suggest that incorporating visual search behavior into transformers may support objective, process-based expertise assessment in radiology. Code is available via \href{https://github.com/leiluk1/GazeToSkill}{https://github.com/leiluk1/GazeToSkill}.

\keywords{Eye tracking \and Skill assessment \and Vision Transformer}

\end{abstract}

\section{Introduction}

Accurate and objective assessment of clinical expertise is essential for developing training programs that promote the adoption of expert-like strategies and accelerate skill acquisition \cite{ashraf2018eye}. In radiology, expertise is reflected in visual search behavior, with consistent differences reported between experts and novices \cite{brunye2019review}. Capturing these differences requires instrumentation capable of tracking spatial and temporal patterns of attention during diagnostic reading, and eye-tracking is one of the tools that enables quantitative measurement of visual behavior \cite{ashraf2018eye,brunye2019review}. However, reported gaze-based differences are often task- and modality-dependent, indicating that handcrafted statistical measures may not fully capture the complexity of visual expertise \cite{gegenfurtner2011expertise,vandergijp2017visual}. 

Deep learning (DL) offers a complementary approach for modeling high-dimensional spatiotemporal gaze patterns, potentially revealing subtle structures associated with expertise \cite{10453951}. Most existing studies dedicated to gaze-based skill assessment typically operate on 2D views or video clips and do not explicitly model the three-dimensional spatial context of volumetric imaging \cite{akerman2023extracting,lam2022machine,pedrett2023technical,sharma2021multimodal}. Therefore, it remains unclear how well these approaches generalize to 3D CT settings. A related line of research within this domain investigates gaze scanpath prediction, where DL models are trained on expert gaze recordings to learn sequential patterns of visual search. After training, these models can generate expert-like gaze trajectories conditioned on input medical images, providing an implicit representation of expert attention strategies, and potentially distinguish experts' scanpaths from novices. However, to the best of our knowledge, only one study has investigated scanpath prediction for 3D CTs \cite{pham2025ct}. 

Motivated by the limited exploration of expertise modeling in 3D medical imaging, we aim to enrich the field of gaze-based skill assessment by investigating expert attention patterns in volumetric CT. We introduce a gaze-informed transformer framework that integrates radiologist visual attention directly into volumetric CT representation learning. Using a DINOv2 vision transformer (ViT), we inject per-patch gaze signals at two complementary levels: (1) within the self-attention mechanism via a learnable log-space bias that multiplicatively reweights attention probabilities, and (2) at the representation level through gaze-weighted pooling of patch embeddings. The additive logit bias increases the influence of highly fixated regions during attention computation while still allowing the transformer to model global context. This design enables the model to learn not only \textit{what} is visible in a CT scan, but also \textit{where} expert radiologists allocate their attention. Furthermore, we collected 182 CT reading sessions with synchronized eye-tracking data from five expert and novice radiologists and compared our methodology against adapted approaches on our dataset.

\section{Dataset}

\subsection{Data Collection}

\noindent \textbf{CT Dataset.} Three expert radiologists (7+ years of professional experience) and two novice readers (<1 year of experience) analyzed 40 lung CT scans sourced from the publicly available LIDC–IDRI dataset \cite{armato2015lidc}. Among the images, 24 scans corresponded to lung cancer patients, while the remaining 16 were from non-cancer patients. The cancer cases contained 47 individual lung nodules, including 20 small-to-intermediate–sized nodules \cite{sanchez2018management} and 27 large nodules \cite{han2017volume}. 

\noindent \textbf{Experimental Protocol.} Gaze data were recorded with a Tobii Eye Tracker 4C (90 Hz) mounted below a 23.7-inch 4K 10-bit monitor, while audio reports were captured via a headset microphone. Radiologists completed calibration before the experiment, with recalibration performed after posture changes. CT scans were reviewed in the RadiAnt DICOM Viewer \cite{radiant}. Recordings were started and ended verbally (“START”/“END”) and synchronized by fixating on a predefined red point while stating “RED POINT.” Radiologists followed their routine clinical workflow, including zooming, adjusting window level (WL) and window width (WW), and navigating across different orthogonal planes.

\begin{figure}[t]
\centering
\includegraphics[width=0.97\textwidth]{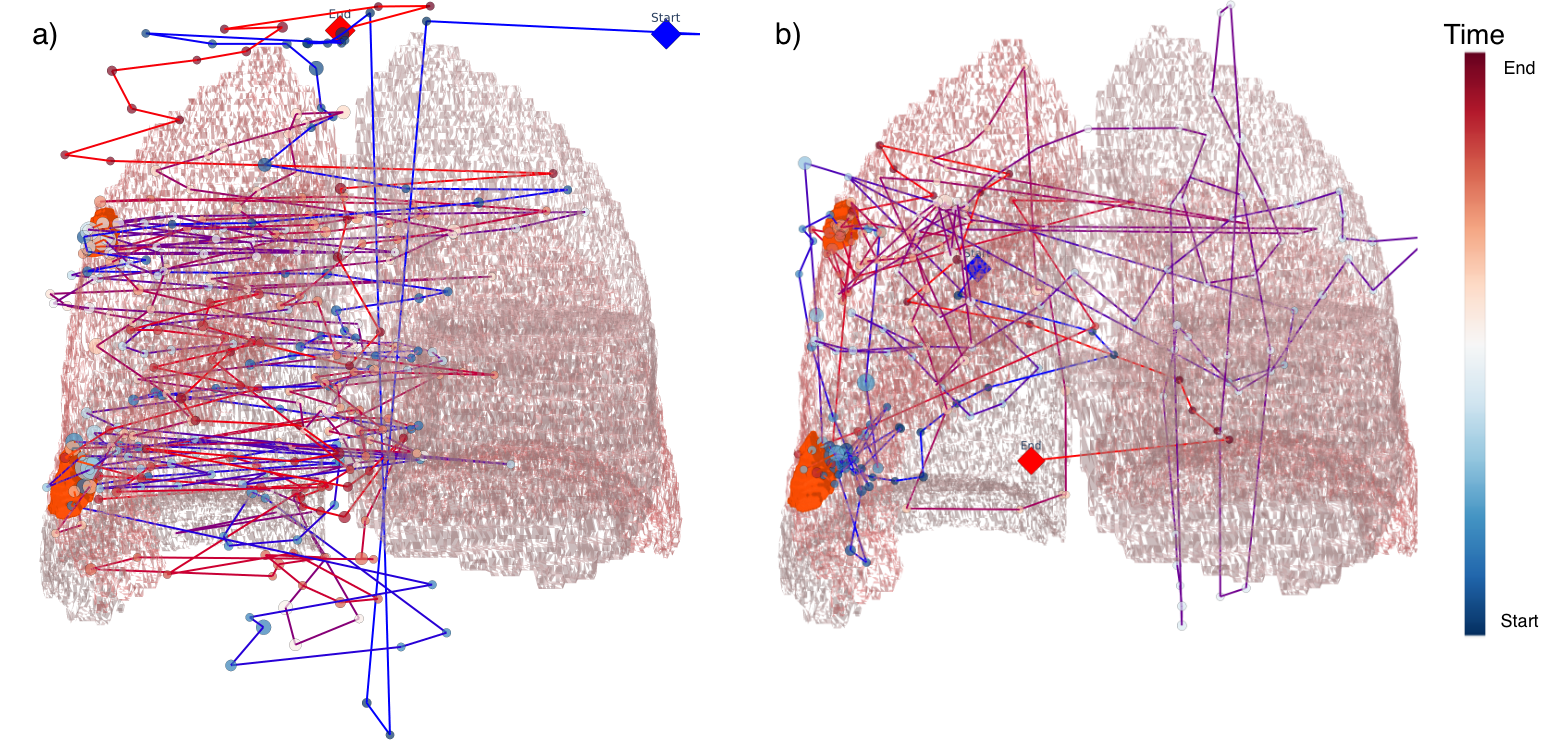}
\caption{Example scanpaths from an expert (a) and a novice (b). The trajectories indicate the sequence of fixations over time, with the time bar normalized. Lung nodules are indicated by \textcolor{nodulecolor}{orange segmentation masks}.} \label{fig1}
\end{figure}

\subsection{Data Preprocessing}

\noindent \textbf{Gaze-to-frame Synchronization.} Raw eye-tracking coordinates were synchronized with video frames, so only frames with gaze data were processed. Gaze drift was corrected by computing the deviation from a reference red point at the start and end of each session and applying time-based linear interpolation between these two data points. In addition, gaze drift was adjusted using the nodule location. For each frame, we used EasyOCR \cite{JaidedAI_EasyOCR} to automatically extract metadata displayed in the viewer, including CT plane, slice number, WL, WW.

\noindent \textbf{Gaze-to-CT Coordinate Mapping.} CT slice numbers were used to retrieve the corresponding slices from the original NIfTI volume and windowed using the extracted WL/WW parameters. Each slice was aligned to the video frame via affine transformation (accounting for scaling, rotation, and translation), and gaze coordinates were mapped from screen space to CT image space using the inverse affine matrix, yielding pixel-accurate locations in the original CT volume.

\noindent \textbf{Fixation Extraction.} Fixations were classified using the Dispersion-Threshold Identification (I-DT) method \cite{Salvucci2000}, which groups spatially clustered gaze points that persist over time. The duration threshold was set to 100 ms and the dispersion threshold to 1° of visual angle, as recommended in the literature \cite{EyeHolmqvist,Salvucci2000}. 

\noindent \textbf{Heatmap Generation.} We generated fixation heatmaps for each slice by smoothing projected gaze points with a Gaussian kernel corresponding to 1° of visual angle. The Gaussian standard deviation was computed per session using the mean viewing distance and the screen-to-CT pixel ratio, enabling subject-specific spatial calibration. The resulting heatmaps were normalized.

\section{Methods}

\begin{figure}[t]
\includegraphics[width=\textwidth]{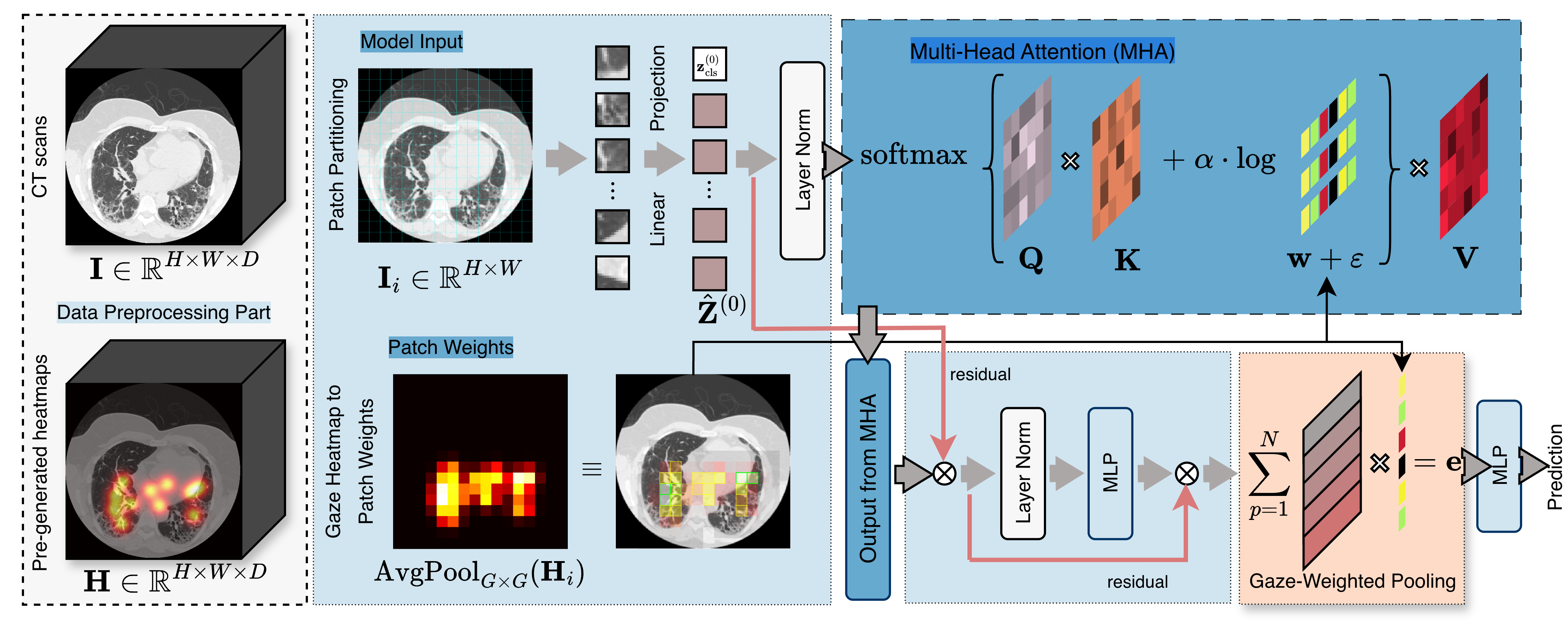}
\caption{Overview of the proposed gaze-informed transformer model. The input CT is partitioned into patches and projected into tokens, while the heatmap volume is downsampled and normalized to obtain per-patch gaze weights. The weights are injected into each self-attention layer as a log-space bias, reweighting attention toward fixated regions. Final slice embeddings are obtained via gaze-weighted pooling, aggregated across slices, and fed to a MLP classifier that predicts the expertise class.} \label{fig1}
\end{figure}

\noindent\textbf{Input.} A CT volume is defined as $\mathbf{I} \in \mathbb{R}^{H \times W \times D}$, where $D$ is a number of axial slices. For each slice $\mathbf{I}_{i} \in \mathbb{R}^{H \times W}$, WW and WL are applied. The greyscale slice is triplicated to form a pseudo-RGB image $\mathbf{X}_i \in \mathbb{R}^{3 \times H' \times W'}$.

\noindent \textbf{Gaze Heatmap to Patch Weights.} Each radiologist's gaze session has a heatmap volume $\mathbf{H} \in \mathbb{R}^{H \times W \times D}$. For slice~$i$, heatmap $\mathbf{H}_i \in \mathbb{R}^{H \times W}$ is downsampled to the patch grid: $\tilde{\mathbf{W}}_i = \mathrm{AvgPool}_{G \times G}(\mathbf{H}_i) \in \mathbb{R}^{G \times G},$ where $G = H'/P$ is the patch grid size and $P$ is the ViT patch size. The pooled map is flattened and $\ell_1$-normalized to obtain per-patch gaze weights $\mathbf{w}$:
\begin{equation}
    \tilde{w}_{i,p} = \mathrm{vec}(\tilde{\mathbf{W}}_i)_p, \quad p = \overline{1, N}, \quad N = G^2.
\end{equation}
\begin{equation}
w_{i,p} =
\begin{cases}
\dfrac{\tilde{w}_{i,p}}{\sum_{j=1}^{N} \tilde{w}_{i,j}} & \text{if } \sum_{j=1}^{N} \tilde{w}_{i,j} > 0, \\[6pt]
\dfrac{1}{N} & \text{otherwise}.
\end{cases}
\end{equation}

\noindent \textbf{Patch Embedding.} Each slice $\mathbf{X}_i \in \mathbb{R}^{3 \times H' \times W'}$ is partitioned into non-overlapping patches of size $P \times P$ and linearly projected to obtain $N$ patch tokens $\mathbf{z}_p^{(0)}, p = \overline{1, N}$. A learnable classification token (CLS) $\mathbf{z}_{\mathrm{cls}}^{(0)} \in \mathbb{R}^{D_e}$, where $D_e$ is embedding dimension, is prepended, and learnable positional embeddings are added. The resulting sequence $\hat{\mathbf{Z}}^{(0)}$ is passed through $L$ transformer blocks.

\noindent \textbf{Gaze-Bias Attention.} At each transformer layer $\ell = \overline{1, L}$,  input $\hat{\mathbf{Z}}^{(\ell-1)}$ is transformed into queries, keys, and values via learned projections $\mathbf{Q}$, $\mathbf{K}$, $\mathbf{V}$. Gaze is injected as an additive log-space bias derived from per-patch gaze weights:
\begin{equation}
    A_{ij}^{\mathrm{gaze}} = \frac{\mathbf{q}_i^\top \mathbf{k}_j}{\sqrt{d_h}} + \alpha \cdot \log(w_j + \varepsilon),
    \label{eq:gaze_bias}
\end{equation}
where $d_h = D_e / n_h$ is the per-head dimension and $n_h$ is the number of attention heads, $\alpha$ is a learnable scalar, $\varepsilon$ is a small constant that prevents numerical underflow, and $w_j$ denotes the gaze weight for key position~$j$. The CLS key receives zero bias by setting $w_0 = 1$. The bias is shared across all heads within a layer and is identical across all $L$~layers. Applying the exponential to the gaze-biased logit yields:
\begin{equation}
\operatorname{softmax}(z_j + \alpha \log w_j) \propto e^{z_j} \cdot e^{\alpha \log w_j} = e^{z_j} \cdot w_j^{\alpha},
\end{equation}
where $z_j = \frac{q_i \cdot k_j}{\sqrt{d_h}}$ denotes the standard attention logit. Thus, the additive log-space bias in logit space becomes a multiplicative scaling in probability space: each key's attention weight is multiplied by $w_j^{\alpha}$, amplifying heavily fixated patches and suppressing those that received little gaze. Then, the output of each transformer block follows the standard residual structure. 

\noindent \textbf{Gaze-Weighted Output Pooling.} Given the final patch tokens $\mathbf{z}_p \in \mathbb{R}^{D_e}$ and normalized gaze weights $w_p$ for slice~$i$, the slice-level embedding is computed as the weighted average:
\begin{equation}
    \mathbf{e}_i = \sum_{p=1}^{N} w_{i,p} \, \mathbf{z}_p.
    \label{eq:gaze_pool}
\end{equation}
This directs the representation towards image regions that received visual attention, encoding where the radiologist looked during interpretation.

\noindent \textbf{Slice Sampling.} During training, a subset of $K$ slices is uniformly sampled from the $D$ available slices. After Gaze-Weighted Output Pooling, the slice-level embeddings are aggregated by mean pooling across all $K$ slices: $\bar{\mathbf{e}} = \frac{1}{K} \sum_{i=1}^{K} \mathbf{e}_i.$

\noindent \textbf{Classification Head.} The session embedding $\bar{\mathbf{e}} \in \mathbb{R}^{D_e}$ is passed through a two-layer MLP:
\begin{equation}
    \hat{\mathbf{y}} = \mathbf{W}_2 \, \mathrm{Dropout}\!\left(\mathrm{ReLU}\!\left(\mathbf{W}_1 \bar{\mathbf{e}} + \mathbf{b}_1\right)\right) + \mathbf{b}_2,
    \label{eq:head}
\end{equation}
where $\mathbf{W}_1 \in \mathbb{R}^{D_h \times D_e}$, $\mathbf{W}_2 \in \mathbb{R}^{2 \times D_h}$, and $D_h$ is the hidden dimension. The output $\hat{\mathbf{y}} \in \mathbb{R}^{2}$ represents logits for the novice and expert classes.

\section{Results}

\noindent \textbf{Dataset.} Our dataset comprises 40 CTs with 8{,}022 axial slices ($\sim$200 slices per volume). Across all radiologists, sessions lasted $117.9 \pm 60.3$\,s and contained $160 \pm 78$ fixations (mean duration $0.48 \pm 0.25$\,s, cumulative dwell time $73.1 \pm 38.0$\,s).

\noindent \textbf{Implementation Details.} After excluding incomplete recordings, 182 samples remained (3 experts: A, B, C and 2 novices: D, E). Radiologists C and E were held out for testing (67 samples: 39 expert, 28 novice), while the remaining 115 samples were used for training. All models were trained using stratified 5-fold cross-validation to address class imbalance ($\sim$70\% expert). At inference phase, fold predictions were averaged, and the final classification threshold was determined using Youden’s J on the test set. For \textit{Skill Assessment Models} as well as for our approach, the training set included 80 expert and 35 novice samples. \textit{Adapted Scanpath Prediction Models} were trained only on expert data (A and B; 80 samples), and expertise classification was based on similarity metrics. 

For training our model, we used Adam with a learning rate of $1e^{-5}$ for the classification head and $1e^{-6}$ for the backbone. Each training step sampled $8$ random slices per CT volume with gradient accumulation over $4$ steps, yielding a batch size of $4$ sessions. Models were trained for 100 epochs with the best checkpoint selected by validation ROC-AUC. The classification head consisted of two linear layers ($768 \rightarrow 256 \rightarrow 2$) with ReLU and dropout (p=$0.1$), trained with binary cross-entropy (BCE) loss. Input CT slices were windowed to lung settings (WL=$-600$, WW=$1500$) and resized to $518 \times 518$. For the DINOv2 ViT-B/14 backbone, the patch size was $14 \times 14$, and the embedding dimension was $768$. The model had 86.8 M trainable parameters and was implemented using PyTorch Lightning v2.6.0 with PyTorch 2.7.0 (CUDA 12.6) and trained on an NVIDIA L40S GPU for approximately 17 hours.

\noindent \textbf{Evaluation.} We compared our approach to two categories of existing models, adapted and trained on our CT data (see Table~\ref{tab:results}). 

\noindent \textit{Skill Assessment Models:} We adapted four multimodal CNN architectures with strong reported performance, originally proposed by Sharma et al.~\cite{sharma2021multimodal} for fetal ultrasound skill classification: Late Fusion (LF-CNN), Intermediate Fusion (IF-CNN), Hybrid Fusion (HF-CNN), and Tensor Fusion (TF-CNN). Original input modalities included Standard Plane (SP), Spatial Gaze Maps (SGP), Gaze Trajectory Images (GTI), and Pupillary Response Images. Pupillary Response Images were omitted due to unavailability; and SP images are operator-acquired anatomical views that reflect skill level, instead we use fixation-ordered CT scans augmented with gaze information (FO-CT+Gaze), where expertise is expressed through slice navigation behavior; generation of SGM and GTI repeated original pipeline \cite{sharma2021multimodal}. We separately tested two modalities (SGM and GTI) and three modalities (FO-CT+Gaze, SGM, GTI) to show contribution of SP replacement. 

\noindent \textit{Adapted Scanpath Prediction Models:} We trained two models on expert gaze data to predict scanpaths; the similarity between a reader's actual scanpath and the model prediction can thus serve as an indicator of expertise. Lou et al. introduced a multi-stream model with three backbones (ConvNeXt-B, HRNet-W48, CSwin Transformer) that achieved state-of-the-art gaze saliency prediction for mammograms \cite{10089554}. We adapted their framework from 2D mammography to chest CT by projecting 3D gaze data into 2D using Maximum Intensity Projection (MIP). We evaluated two MIP strategies: (1) a session-level MIP, collapsing all viewed slices per reading into a single saliency map, and (2) a chunk-level MIP, splitting each reading into groups of 20 consecutive slices to preserve local depth context. We then thresholded the skill prediction based on Normalized Scanpath Saliency (NSS) metric. CT-Searcher~\cite{pham2025ct}, a transformer-based scanpath predictor, was trained on the CTScanGaze dataset \cite{pham2025ct} and fine-tuned on our expert data. The predictions were then thresholded based on MultiMatch vector similarity~\cite{dewhurst2012depends}.

\begin{table}[t]
\centering
\caption{Comparison against prior methods on the held-out test set using ROC-AUC, F1, Sensitivity, and Specificity metrics. SGP: Spatial Gaze Map; GTI: Gaze Trajectory Image; FO-CT+Gaze: fixation-ordered CT slices augmented with gaze information.}
\label{tab:results}
\setlength{\tabcolsep}{3.5pt}
\scriptsize
\renewcommand{\arraystretch}{1.5}
\begin{tabular}{lllcccc}
\hline
Model & Input & Prediction & ROC-AUC & F1 & Sens. & Spec. \\ \hline
\multirow{2}{*}{TF-CNN \cite{sharma2021multimodal}} & SGP, GTI & \multirow{2}{*}{Skill} & 0.7793 & 0.8372 & \textbf{0.9231} & 0.6071 \\
& FO-CT+Gaze, SGP,  GTI & & 0.7454 & 0.7568 & 0.7179 & 0.7500 \\
\arrayrulecolor{lightgray}\cline{1-7}
\arrayrulecolor{black}

\multirow{2}{*}{IF-CNN \cite{sharma2021multimodal}} & SGP, GTI & \multirow{2}{*}{Skill} & 0.7308 & 0.6866 & 0.5897 & 0.8214 \\
& FO-CT+Gaze, SGP,  GTI & & 0.7363 & 0.7532 & 0.7436 & 0.6786 \\
\arrayrulecolor{lightgray}\cline{1-7}
\arrayrulecolor{black}

\multirow{2}{*}{LF-CNN \cite{sharma2021multimodal}} & SGP, GTI & \multirow{2}{*}{Skill} & 0.7463 & 0.7838 & 0.7436 & 0.7857 \\
& FO-CT+Gaze, SGP,  GTI & & 0.7637 & 0.8205 & 0.8205 & 0.7500 \\

\arrayrulecolor{lightgray}\cline{1-7}
\arrayrulecolor{black}

\multirow{2}{*}{HF-CNN \cite{sharma2021multimodal}} & SGP, GTI & \multirow{2}{*}{Skill} & 0.7518 & 0.7838 & 0.7436 & 0.7857 \\
& FO-CT+Gaze, SGP,  GTI & & 0.7363 & 0.7733 & 0.7436 & 0.7500 \\

\arrayrulecolor{black}\cline{1-7}
\arrayrulecolor{black}


\multirow{2}{*}{Lou et al. \cite{10089554}} & FO-CT session-level  & \multirow{2}{*}{Saliency map} & 0.6782 & 0.7123 & 0.6667 & 0.7241 \\
& FO-CT chunk-level & & 0.5570 & 0.6000 & 0.5385 & 0.6552 \\
\arrayrulecolor{lightgray}\cline{1-7}

CT-Searcher \cite{pham2025ct} & CT volume & Scanpath & 0.8750 & 0.8421 & 0.8205 & 0.8214 \\ 

\arrayrulecolor{black}\cline{1-7}
\arrayrulecolor{black}

\textbf{Ours} & CT volume & Skill & \textbf{0.9089} & \textbf{0.8611} & 0.7949 & \textbf{0.9310} \\ 

\hline
\end{tabular}
\end{table}

\noindent \textbf{Ablation study.} We evaluated DINOv2 \cite{oquab2023dinov2}, Med3D \cite{chen2019med3d}, SwinUNETR \cite{10.1007/978-3-031-43901-8_40}, and UniMISS \cite{xie2022unimiss} for CT slice embeddings. Med3D \cite{chen2019med3d} showed representational collapse (cross-CT 0.9998; variance 0.0002), indicating minimal discriminative capacity; UniMISS \cite{xie2022unimiss} demonstrated weak inter-patient separation ($0.9801 \pm 0.0193$). SwinUNETR \cite{10.1007/978-3-031-43901-8_40} improved between-patient discrimination ($0.9349 \pm 0.0668$) but exhibited near-saturated consecutive similarity (0.9999), suggesting limited sensitivity to subtle slice-level variations. DINOv2 \cite{oquab2023dinov2} achieved the best trade-off, maintaining clear inter-patient separation ($0.9381 \pm 0.0261$) with higher representational variance ($0.6814$), and was therefore selected as a backbone. 

To assess individual contributions of gaze-integration mechanisms, we conducted an ablation study (Table \ref{tab:ablat}). Firstly, we trained MLP with frozen DINOv2 embeddings and obtained near-random results for prediction (Mean Val AUC: $0.4455 \pm 0.045$, AUC on test: $0.5234$). Then, we unfroze DINOv2 and tested configurations over two axes: the attention mode (Gaze-Bias, Fixation-Mask, or None) and pooling strategy (Gaze-Weighted or CLS). 

\section{Discussion and Conclusion}

According to Table~\ref{tab:results}, our proposed model achieved the best performance across almost all metrics, with the highest AUC of $0.91$, F1 of $0.86$, and specificity of $0.93$. The multi-modal fusion architectures introduced by Sharma et al. \cite{sharma2021multimodal}, originally designed for ultrasound, achieved comparable performance when adapted to CT, indicating that gaze patterns reflect general expertise rather than task-specific behaviors. Among these, TF-CNN with SGP, GTI inputs achieved the highest sensitivity ($0.92$), though with lower specificity ($0.61$), indicating that novices were frequently misclassified as experts. The addition of CT image information (FO-CT+Gaze) did not consistently improve performance across architectures, with IF-CNN showing little improvement (ROC-AUC: $0.73 \rightarrow 0.74$) and LF-CNN (ROC-AUC: $0.75 \rightarrow 0.76$). This suggests that the FO-CT+Gaze representation may provide limited additional value beyond other features, or that the increased model complexity leads to overfitting given the dataset size. Lou et al. \cite{10089554} relies on a three-encoder architecture that is computationally expensive in both time and memory, limiting its scalability to volumetric data. Chunk-level processing, necessary to reduce computational demands, achieved the ROC-AUC of only $0.56$, suggesting that this architecture may be difficult to adapt fully for 3D CT interpretation tasks. CT-Searcher \cite{pham2025ct}, originally designed for scanpath task, achieved accurate results (ROC-AUC $0.88$) despite not being explicitly designed for skill classification, indicating that learned representations for scanpath modeling can encode skill-relevant information. 

\begin{table}[t]
\centering
\scriptsize
\caption{Attention variants: Gaze-Bias - soft additive bias injected into self-attention layers; Fix-Mask - hard binary mask on attention, meaning patches with zero gaze weight are blocked from being attended to; None - standard DINOv2 self-attention, no gaze signal enters the transformer. Pooling strategies: Gaze-Weighted - weighted average of all patch tokens using normalized gaze weights, i.e., patches the radiologist looked at more contribute more to the session embedding; CLS - uses only CLS token.}
\label{tab:ablat}
{\setlength{\tabcolsep}{4pt}
\begin{tabular}{llllllll}
\hline
Attention & Pooling strategies & Mean Val AUC & AUC & F1 & Sens. & Spec. \\ \hline
Gaze-Bias & Gaze-Weighted & 0.796 $\pm$ 0.091 & \textbf{0.909} & \textbf{0.861} & \textbf{0.795} & 0.931 \\
Gaze-Bias & CLS & 0.828 $\pm$ 0.101 & 0.897 & 0.824 & 0.751 & 0.892 \\
None & Gaze-Weighted & 0.632 $\pm$ 0.090 & 0.856 & 0.759 & 0.623 & \textbf{0.969} \\
Fix-Mask & Gaze-Weighted & 0.871 $\pm$ 0.101 & 0.842 & 0.783 & 0.688 & 0.892 \\ 
\hline
\end{tabular}}
\end{table}

The ablation study, provided in Table \ref{tab:ablat}, reveals that gaze information contributes to expertise classification through two complementary mechanisms. The best-performing configuration (Gaze-Bias + Gaze-weighted, AUC = $0.91$) integrates gaze at both the attention and pooling levels, substantially outperforming variants where gaze operates at only one stage: either biasing attention alone (Gaze-Bias + CLS, AUC = $0.89$) or weighting pooling alone (None + Gaze-weighted, AUC = $0.86$). This suggests that gaze serves two distinct roles: during feature extraction, it guides the model to focus its internal processing on regions the radiologist actually examined, while during aggregation, it ensures that heavily fixated areas contribute more to the final classification. In other words, gaze tells the model both how to analyze each slice and what matters most within it, and both signals are needed to distinguish experts from novices.

The clinical significance of our work lies in enabling objective image reading assessment of radiologist expertise. This has practical implications for radiology training, where automated gaze-based feedback could identify specific visual weaknesses in trainees. Understanding which gaze behaviors reliably mark expertise may also inform the design of computer-aided detection systems tailored to the perceptual gaps of less experienced readers. Future work will focus on increasing the number of participants from five radiologists to support more extensive clinical validation.

\bibliographystyle{splncs04}
\bibliography{report} 

\end{document}